\documentclass[conference]{IEEEtran}
\usepackage{amsmath}
\usepackage{graphicx}  %%% for including graphic
\usepackage{url}
\usepackage{cite}

\ifCLASSINFOpdf
\else
\fi
\renewcommand\IEEEkeywordsname{Keywords}

\begin{document}
%\title{Prediction of Energy Expenditure by Machine Learning Analysis of Heart Activity after Exercises}

\title{Prediction of Physical Load Level by Machine Learning Analysis of Heart Activity after Exercises}

\author{
\IEEEauthorblockN{Peng Gang$^1$, Wei Zeng$^2$}
\IEEEauthorblockA{School of Information Science and Technology\\ Huizhou University\\
Huizhou, China\\
Email: $^1$peng@hzu.edu.cn\\ $^2$weizeng.hzu@yahoo.com}
\and
\IEEEauthorblockN{Yuri Gordienko$^3$, Oleksandr Rokovyi,\\ Oleg Alienin,  and Sergii Stirenko$^4$}
\IEEEauthorblockA{National Technical University o Ukraine\\ ``Igor Sikorsky Kyiv Polytechnic Institute''\\
Kyiv, Ukraine\\
Email: $^3$yuri.gordienko@gmail.com\\ $^4$sergii.stirenko@gmail.com}
}


%Telephone: (800) 555--1212, Fax: (888) 555--1212}


\maketitle

\begin{abstract}
The assessment of energy expenditure in real life is of great importance for monitoring the current physical state of people, especially in work, sport, elderly care, health care, and everyday life even. 
This work reports about application of some machine learning methods (linear regression, linear discriminant analysis, k-nearest neighbors, decision tree, random forest, Gaussian naive Bayes, support-vector machine) for monitoring energy expenditures in athletes. The  classification problem was to predict the known level of the in-exercise loads (in three categories by calories) by the heart rate activity features measured during the short period of time (1 minute only) after training, i.e by features of the post-exercise load.
The results obtained shown that the post-exercise heart activity features preserve the information of the in-exercise training loads and allow us to predict their actual in-exercise levels. The best performance can be obtained by the random forest classifier with all 8 heart rate features (micro-averaged area under curve value $AUC_{micro}=0.87$ and macro-averaged one $AUC_{macro}=0.88$) and the k-nearest neighbors classifier with 4 most important heart rate features ($AUC_{micro}=0.91$ and $AUC_{macro}=0.89$).
The limitations and perspectives of the ML methods used are outlined, and some practical advices are proposed as to their improvement and implementation for the better prediction of in-exercise energy expenditures.
\end{abstract}


% \fi
\IEEEpeerreviewmaketitle

\section{Introduction}
The everyday physical activity has a great impact on a  health of people. The assessment of energy expenditure in real life is of great importance for monitoring the current physical state of people, especially in work, sport, elderly care, and everyday life even. Numerous researches demonstrated that inactive behavior can cause serious health-related problems, and decay to death even. That is why the assessment of energy expenditure in real life has become very important \cite{Halson_2014}. In addition highly physically active people (like professional athletes and amateurs even) are interested to improve their performance and increase their training load, for example by increase of frequency, duration, and intensity of trainings. In this context, monitoring the training loads (and related energy expenditures) is very important to determine whether a person is adapting to the training program and to minimize the risk of overtraining that can lead to injury, illness, and death even \cite{Halson_2014}. Unfortunately, despite the active research in this area and thorough monitoring among professional athletes, the most part of such monitoring data and the related conclusions is not available, because it is private and unpublished.

The main aim of this article is to report our results of investigation of available machine learning tools for monitoring energy expenditures in athletes and to propose some practical advices as to their implementation for prediction of energy expenditures after exercises. The section \emph{\ref{Background}.Background and Related Work} gives the very short outline of the current attempts to monitor and estimate training loads. The section \emph{\ref{Methodology}.Methodology} describes the machine learning (ML) methods and dataset used here for prediction of physical load level by ML analysis of heart activity after exercises. The section \emph{\ref{Results}.Results} contains the output results obtained as to the prediction of physical load level by ML analysis of the all available data gathered during exercises and the heart activity data after exercises. The section \emph{\ref{Discussion}.Discussion} is dedicated to discussion of the results obtained and future work planned.



%\hfill September 17, 2014

\section{Background and Related Work}\label{Background}
Monitoring training load can be organized by external or internal means. External load is the work performed by a person without taking into account the person's internal state. And internal load is the relative physiological and psychological stress caused by the external load. The external and internal loads are very important for training monitoring, because the relationship between them can be crucial for estimation of fresh and fatigued states of athletes \cite{Halson_2014,Wallace_2014}. For example, the same power output may be maintained for the same duration of training. But depending on the fatigue state of the athlete, this may be achieved with the higher/lower heart rates, lactate levels, etc \cite{Halson_2014,Taylor_2012}.

Now a number of commercial technologies are available for monitoring the external and internal training loads. Various devices can measure power output during cycling \cite{Jobson_2009}. The training data can be recorded, collected and analyzed to provide information on a number of parameters, including average power, normalized power, speed, and accelerations. The various metrics are derived on the basis of conversion of the cycling power output, for example Training Stress Score$^{TM}$(TSS$^{TM}$) by Training Peaks \cite{Halson_2014,Wallace_2014}. The time-motion analysis (TMA) uses global positioning system (GPS) and digital video to monitor athletes \cite{Halson_2014,Taylor_2012}. In addition to them, isokinetic and isoinertial dynamometry are often utilized to use neuromuscular activity to estimate jump and sprint performance, mean power, peak velocity, peak force, jump height, flight time, contact time, and rate of force development. Unfortunately, most of these methods require specialized equipment with proprietary data processing techniques \cite{Halson_2014,Twist_2013}.

At the same time various internal physiological and perceptual indicators, like heart rate (HR) \cite{Bannister_1980,Edwards_1993,Manzi_2009}, blood lactate (BL) \cite{Beneke_2011}, maximal oxygen uptake (VO$_{2max}$) \cite{Chen_2002,Portier_2001}, and others including different biochemical (steroid, peptide and immune) measures \cite{Papacosta_2011}, are used to get markers of the internal load experienced by the athlete and obtain information as to the current state, fatigue, and recovery of the person.  

The various HR variability (HRV) parameters and HRV-related methodological approaches were intensively used recently \cite{Halson_2014,Plews_2013,Bellenger_2016,Dong_2016,Javaloyes_2019,Vesterinen_2016,Singh_2018_part1,Singh_2018_part2}. Despite the significant success in the instant estimation of training load, there are some open questions as to applicability of the training load estimations for estimation of the related fatigue and recovery. That is why investigations of the potential correlation between the training internal loads during training (in-exercise) and after training (post-exercise) are necessary. For this purpose in this work the measurements of in-exercise or post-exercise HRV are used. The HRV parameters are based on the beat-to-beat (NN or RR) intervals \cite{Shaffer_2017}:
\begin{itemize}
\item AVNN --- the mean value for all NN intervals;
\item SDNN --- the standard deviation value for all NN intervals;
\item RMSSD --- the root mean square value of successive differences between the nearest NNs;
\item SDSD --- the standard deviation value of successive differences between the nearest NNs;
\item NN50 --- the number of pairs of successive NNs that differ by more than 50 ms;
\item pNN50 --- the proportion of NN50, i.e. the NN50 value divided by the total number of NNs;
\item HRV index --- the integral of the density of the NN interval histogram divided by its height. 
\end{itemize}

At the same time, recently progress of machine learning (ML) and deep learning (DL) methods allowed to process effectively the complex datasets with numerous characteristics (features) and find some hidden patterns and trends among the data. For example, several statistical and machine learning methods were proposed recently to classify the type and estimate the intensity of physical load and accumulated fatigue. The data for various types and levels of physical activities and for people with various physical conditions were collected by the wearable heart monitor \cite{Stirenko_2018_Parallel}. That is why some of the well-known ML methods were applied here to find the potential correlation between the training internal loads during training (in-exercise load) and after training (post-exercise load).

\section{Methodology}\label{Methodology}
The problem was formulated as a classification task where the known level of the in-exercise load should be predicted by the features of the athlete physical state after training, i.e by features of the post-exercise load. The levels of in-exercise loads were determined on the basis of the energy expenditures (in calories) calculated and reported by the commercial proprietary algorithm based on the neural network as it was stated in the relevent product description and publications (see below). The algorithm is used in their widely available gadgets proposed for professional and amateur athletes on the basis of heart rate variability (HRV) studies and related findings \cite{Firstbeat_2015, Rennie_2001, Hiilloskorpi_2003, Pulkkinen_2005, Vesterinen_2013}. Actually the following three in-exercise training load levels (in calories) were selected: low (0-400), medium (400-1000), high (1000-4000).    

\subsection{Machine Learning Methods}
The following ML methods were used in the work: 
Linear Regression (LR) \cite{Galton_1886}, Linear Discriminant Analysis (LDA) \cite{Fisher_1936, Rao_1948}, K-Nearest Neighbors (KN) \cite{Cover_1967}\cite{Rao_1948}, Decision Tree (DT) \cite{Breiman_1984}, Random Forest (RF) \cite{Ho_1995}, Gaussian Naive Bayes (GNB) \cite{Zhang_2004}, Support-Vector Machine (SVM) \cite{Ben_2001}.
As to the details of their implementation, LR, DT and RF were applied with the random seed and weighted classes (weights were inversely proportional to class frequencies), LDA --- with singular value decomposition solver, KN --- with the weights adjusted inversely of their distance.

\subsection{Dataset}
The data for these in-exercise training load levels were collected for more than one year history of the training ($>300$ training records) of the well-trained triathlete. They included the following in-exercise load parameters: type of exercise (swimming, cycling, running), distance (from 300 meters for swimming to 182 kilometers for cycling), duration (from 10 minutes for swimming to 9 hours for cycling), instant HR (from 45 to 195 beats per minute), average HR (AHR), maximum HR (MHR), calories, cadence, vertical oscillations, etc. The data for the corresponding post-exercise training load levels were collected by 1-minute measurement of the heart activity after completion of the exercise  (after 30 minutes, but not later than 1 hour). The following features of the post-exercise heart activity were used: AVNN, SDNN, RMSSD, NN50, pNN50, HRV, RAHR (post-exercise AHR), RMHR (post-exercise MHR).

\section{Results}\label{Results}
To investigate the feasibility of the before mentioned ML methods for prediction of physical load level the two sets of experiments were carried out:
\begin{itemize}
\item in-exercise prediction --- classification of training load levels (in aforementioned ranges of calories) by the parameters recorded \emph{during} the training for the various durations (from 15 minutes to 9 hours);
\item post-exercise prediction --- classification of training load levels (in aforementioned ranges of calories) by heart activity parameters recorded \emph{after} the training during 1 minute only. 
\end{itemize}

\subsection{In-exercise Prediction}
For in-exercise classification of the training load levels by calories ($C$) the following model (\ref{eq_in_model_full}) was used with the parameters recorded during the training: type of activity ($A$), distance ($D$), duration ($T$), $AHR$, $MHR$, impulse ($I=T*AHR$), velocity ($V=D/T$), power ($P=V*V$).
\begin{equation}
%	Calories \tilde type+distance+duration+AHR+MHR+impulse+velocity+power 
	C \sim A+D+T+AHR+MHR+I+V+P 
	\label{eq_in_model_full}
\end{equation}

\begin{figure}[!hb]
    \centering
    \includegraphics[trim={1cm 1cm 1cm 2cm},clip,width=8cm]{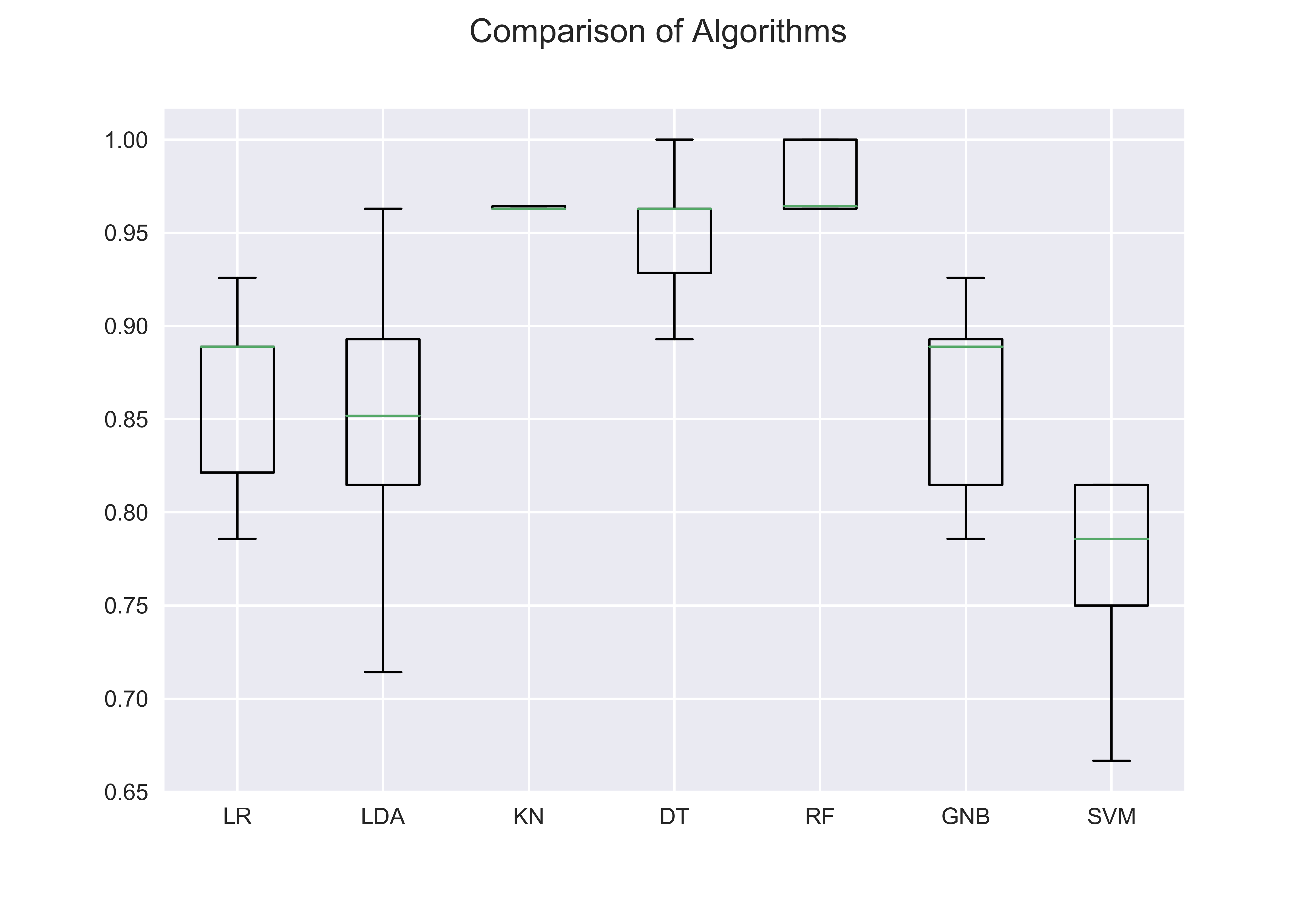}
    \caption{Comparison of ML methods after K-fold cross-validation for \emph{in}-exercise prediction.}
    \label{fig1_IN_01_model_comparison}
\end{figure}

The results of K-fold cross-validation for all before mentioned models are presented in Fig.\ref{fig1_IN_01_model_comparison} and Table \ref{table1_IN_01_model_comparison}. 

\begin{table}[!hb]
    \caption{Accuracy of ML methods for in-exercise prediction.}
    \label{table1_IN_01_model_comparison}
    \centering
    \begin{tabular}{|c||c|c|c|c|c|c|c|}
        \hline
        Method & LR & LDA & KN & DT & RF & GNB & SVM\\
        \hline
        Mean & 0.86 & 0.85 & 0.963 & 0.95 & 0.98 & 0.86 & 0.80\\
        \hline
        Std & 0.05 & 0.08 & 0.001 & 0.04 & 0.02 & 0.05 & 0.10\\
        \hline
    \end{tabular}
\end{table}

\begin{figure}[!hbt]
    \centering
    \includegraphics[trim={1cm 0cm 1cm 1.5cm},clip,width=8cm]{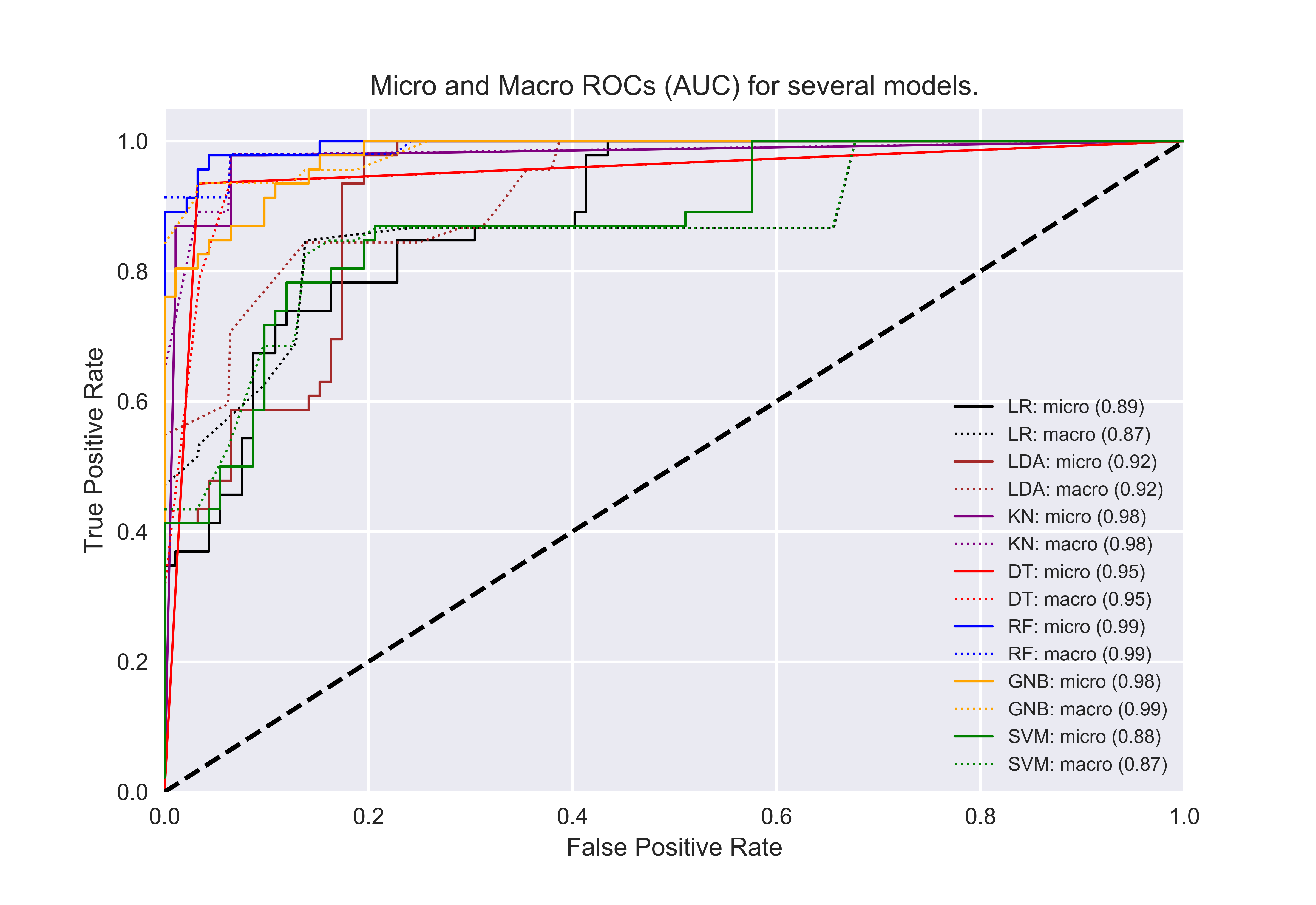}
    \caption{Receiver operating characteristic (ROC) curves and area under ROC-curves (AUC) for \emph{in}-exercise prediction for all models.}
    \label{fig2_IN_06_ROC_model_comparison}
\end{figure}

The corresponding receiver operating characteristic (ROC) curves and area under ROC-curves (AUC) for all before mentioned ML methods and test data (25\% of  the whole dataset) are presented in Fig.\ref{fig2_IN_06_ROC_model_comparison}. A macro-average ROC (``macro'' label in the legend of Fig.\ref{fig2_IN_06_ROC_model_comparison}) takes the average of the precision and recall of the system on different classes and then take the mean value (i.e. treating all classes equally). A micro-average ROC (``micro'' label in the legend of Fig.\ref{fig2_IN_06_ROC_model_comparison}) collects the individual true positives, false positives, and false negatives for different classes and then use them to compute the metrics.

\begin{figure}[!hbt]
    \centering
    \includegraphics[trim={1cm 0cm 1cm 1.5cm},clip,width=8cm]{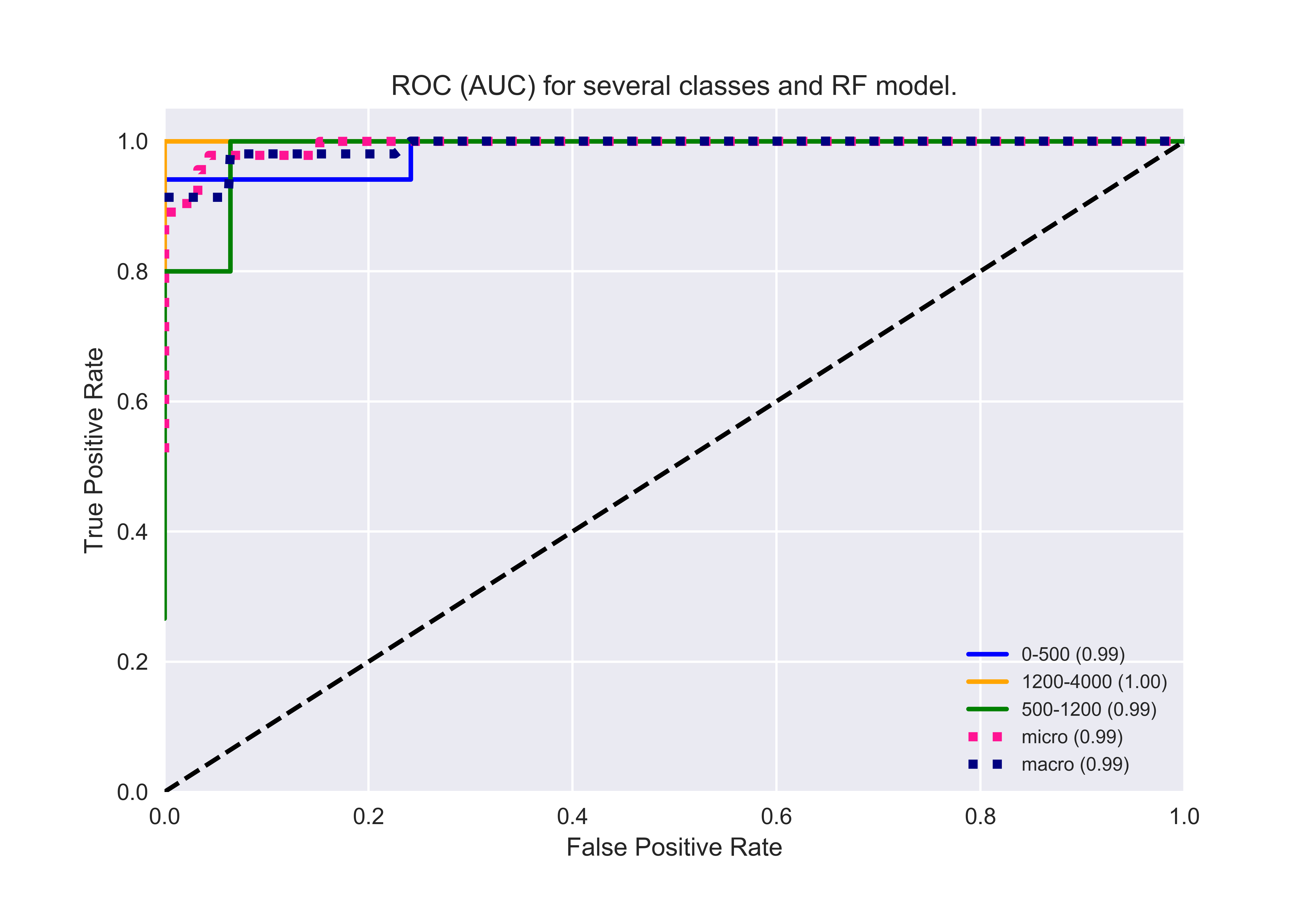}
    \caption{Receiver operating characteristic (ROC) curves and area under ROC-curves (AUC) for \emph{in}-exercise prediction for RF method (for separate classes and for the whole dataset).}
    \label{fig3_IN_05_ROC_several_classes_RF}
\end{figure}

The higher performance can be obtained by random forest classifier (RF), k-nearest neighbors (KN), decision tree (DT), Gaussian naive Bayes (GNB), but not by support-vector machine classifier (SVM), linear discriminant analysis (LDA), and linear regression (LR).
The best performance is demonstrated by the RF classifier (micro-averaged area under curve value $AUC_{micro}=0.99$ and macro-averaged one $AUC_{macro}=0.99$) and the GNB classifier ($AUC_{micro}=0.99$ and $AUC_{macro}=0.98$).
The correspondent ROC-curves with the AUC-values for different classes (levels of training load in calories) and for the whole dataset are shown for RF in Fig.\ref{fig3_IN_05_ROC_several_classes_RF}. 

\begin{figure}[!hbt]
    \centering
    \includegraphics[trim={1cm 2cm 2cm 2cm},clip,width=7cm]{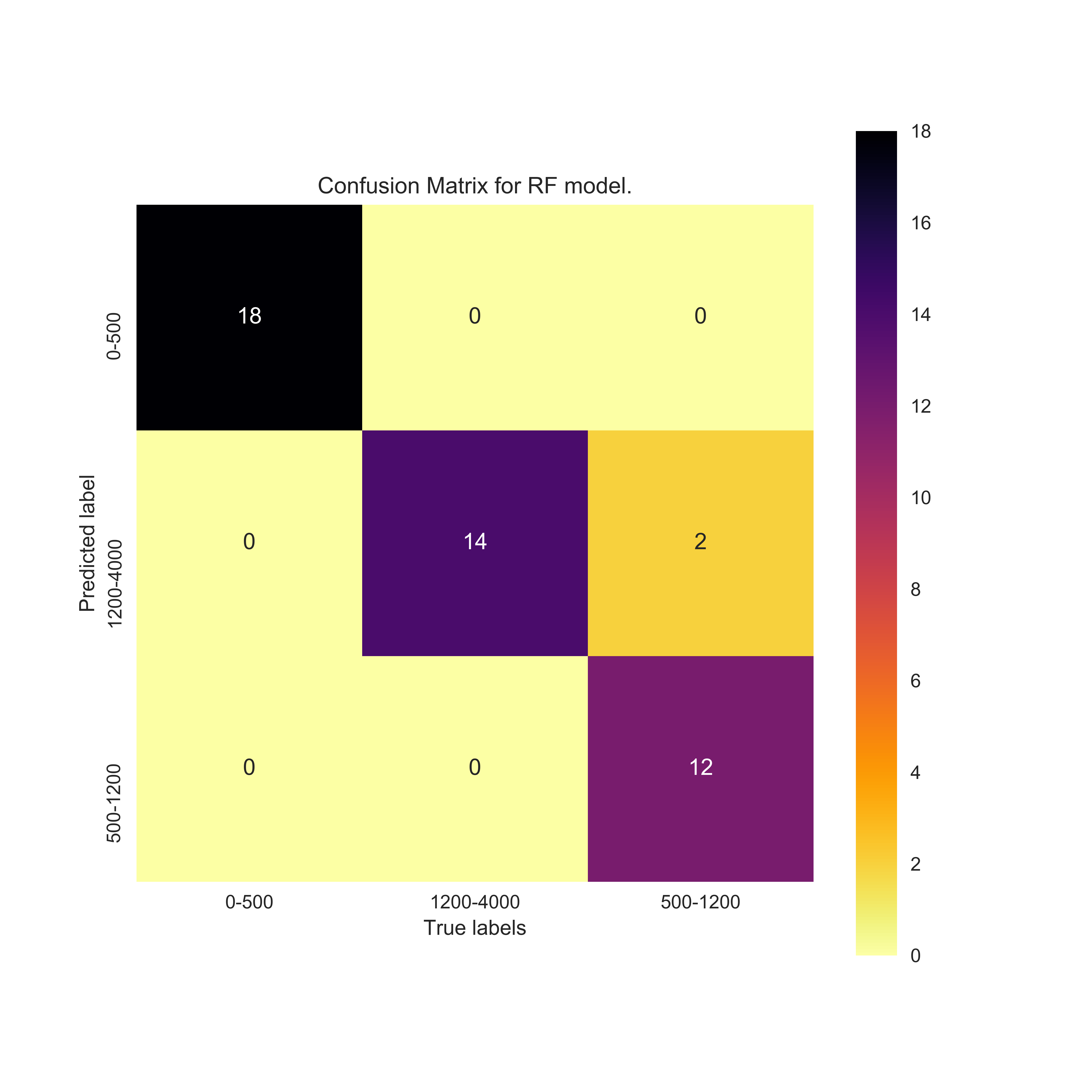}
    \caption{Confusion matrix for \emph{in}-exercise prediction by RF method.}
    \label{fig4_IN_03_confusion_matrix_RF}
\end{figure}

The confusion matrix for prediction of training load levels by RF method (in three aforementioned ranges of calories) by the parameters recorded during the training for the test data (25\% of the whole dataset) is shown on Fig.\ref{fig4_IN_03_confusion_matrix_RF}.

\subsection{Post-exercise Prediction}
Similarly, for post-exercise classification of the training load levels by calories ($C$) the following model (\ref{eq_post_model_full}) was used with heart activity parameters recorded after the training: type of activity ($A$), $AVNN$, $SDNN$, $RMSSD$, $NN50$, $pNN50$, $HRV$, $RAHR$, $RMHR$.
\begin{equation}
    \begin{split}
	    C \sim A+AVNN+SDNN+RMSSD+ \\ +NN50+pNN50+HRV+RAHR+RMHR 
    \end{split}	
	\label{eq_post_model_full}
\end{equation}

\begin{figure}[!hb]
    \centering
    \includegraphics[trim={1cm 1cm 1cm 2cm},clip,width=8cm]{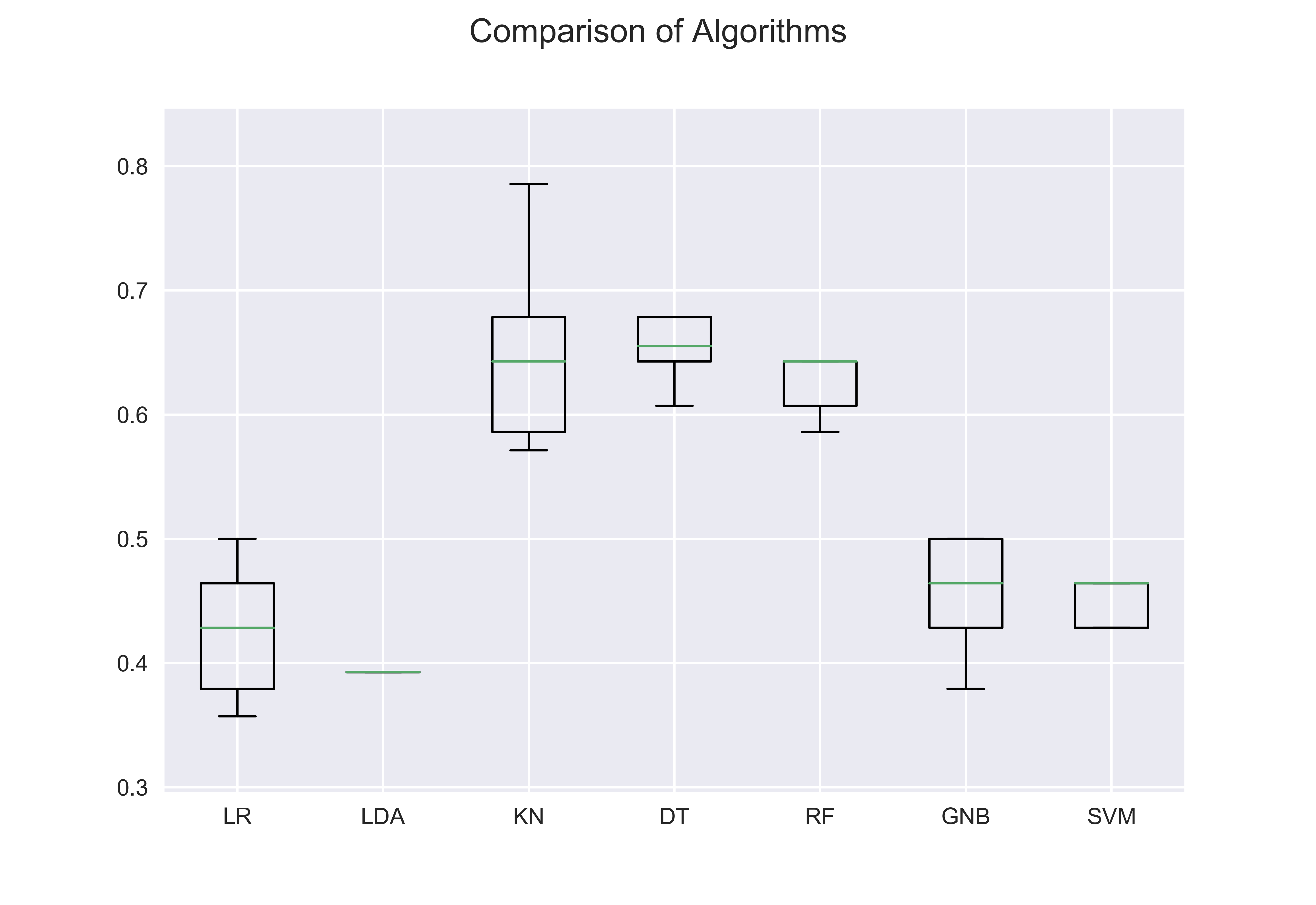}
    \caption{Comparison of ML methods after K-fold cross-validation for \emph{post}-exercise prediction.}
    \label{fig5_POST_01_model_comparison}
\end{figure}

The results of K-fold cross-validation for all before mentioned models are presented in Fig.\ref{fig5_POST_01_model_comparison} and Table \ref{table2_POST_01_model_comparison}. 

\begin{table}[!hb]
    \caption{Accuracy of ML methods for post-exercise prediction.}
    \label{table2_POST_01_model_comparison}
    \centering
    \begin{tabular}{|c||c|c|c|c|c|c|c|}
        \hline
        Method & LR & LDA & KN & DT & RF & GNB & SVM\\
        \hline
        Mean & 0.43 & 0.38 & 0.65 & 0.68 & 0.65 & 0.45 & 0.45\\
        \hline
        Std & 0.05 & 0.03 & 0.08 & 0.07 & 0.07 & 0.05 & 0.06\\
        \hline
    \end{tabular}
\end{table}

\begin{figure}[!hbt]
    \centering
    \includegraphics[trim={1cm 0cm 1cm 1.5cm},clip,width=8cm]{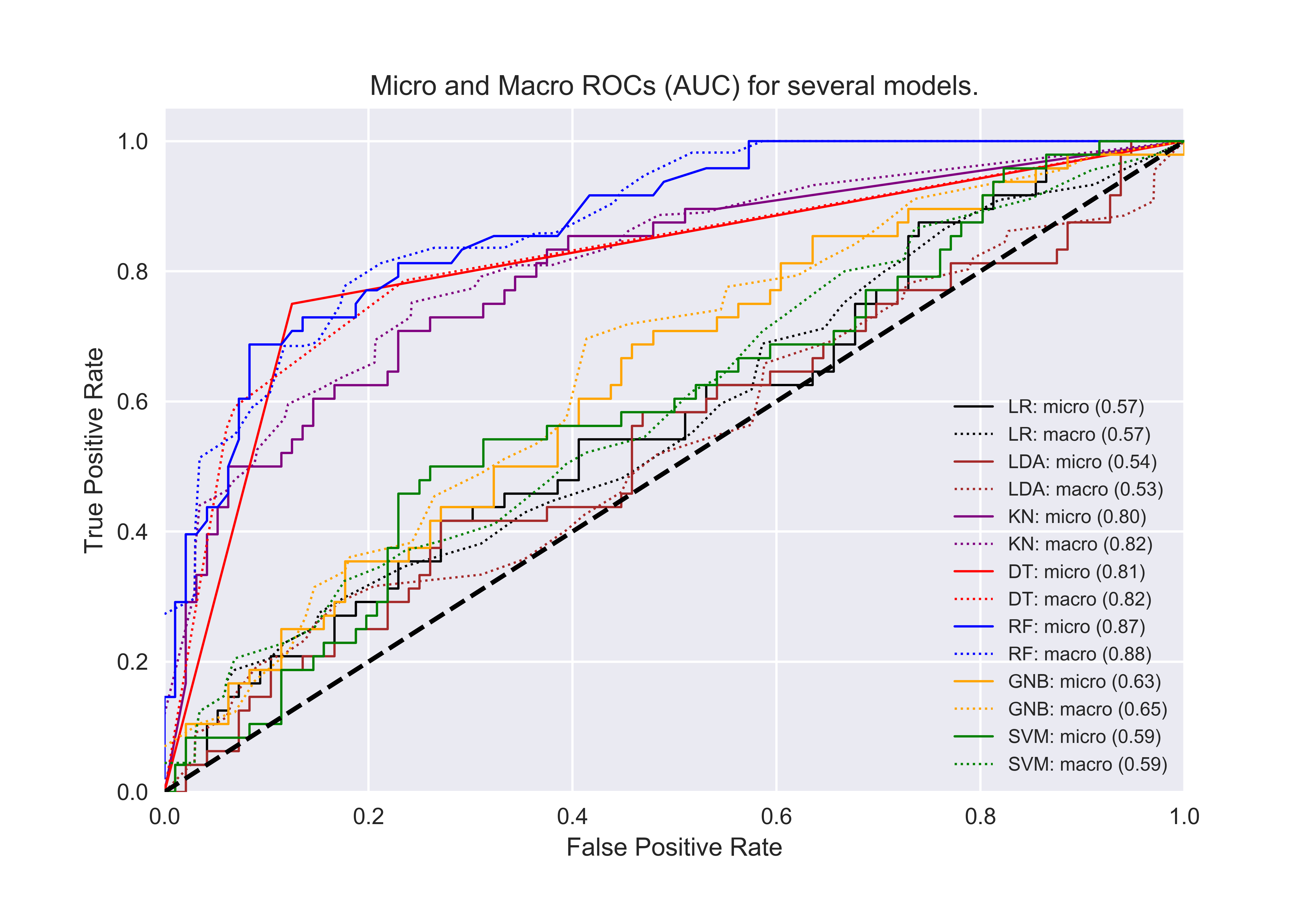}
    \caption{Receiver operating characteristic (ROC) curves and area under ROC-curves (AUC) for \emph{post}-exercise prediction for all models.}
    \label{fig6_POST_06_ROC_model_comparison}
\end{figure}

The corresponding ROC-curves and AUC-values for all aforementioned ML methods and test data (25\% of  the whole dataset) are presented in Fig.\ref{fig6_POST_06_ROC_model_comparison}.

\begin{figure}[!hbt]
    \centering
    \includegraphics[trim={1cm 0cm 1cm 1.5cm},clip,width=8cm]{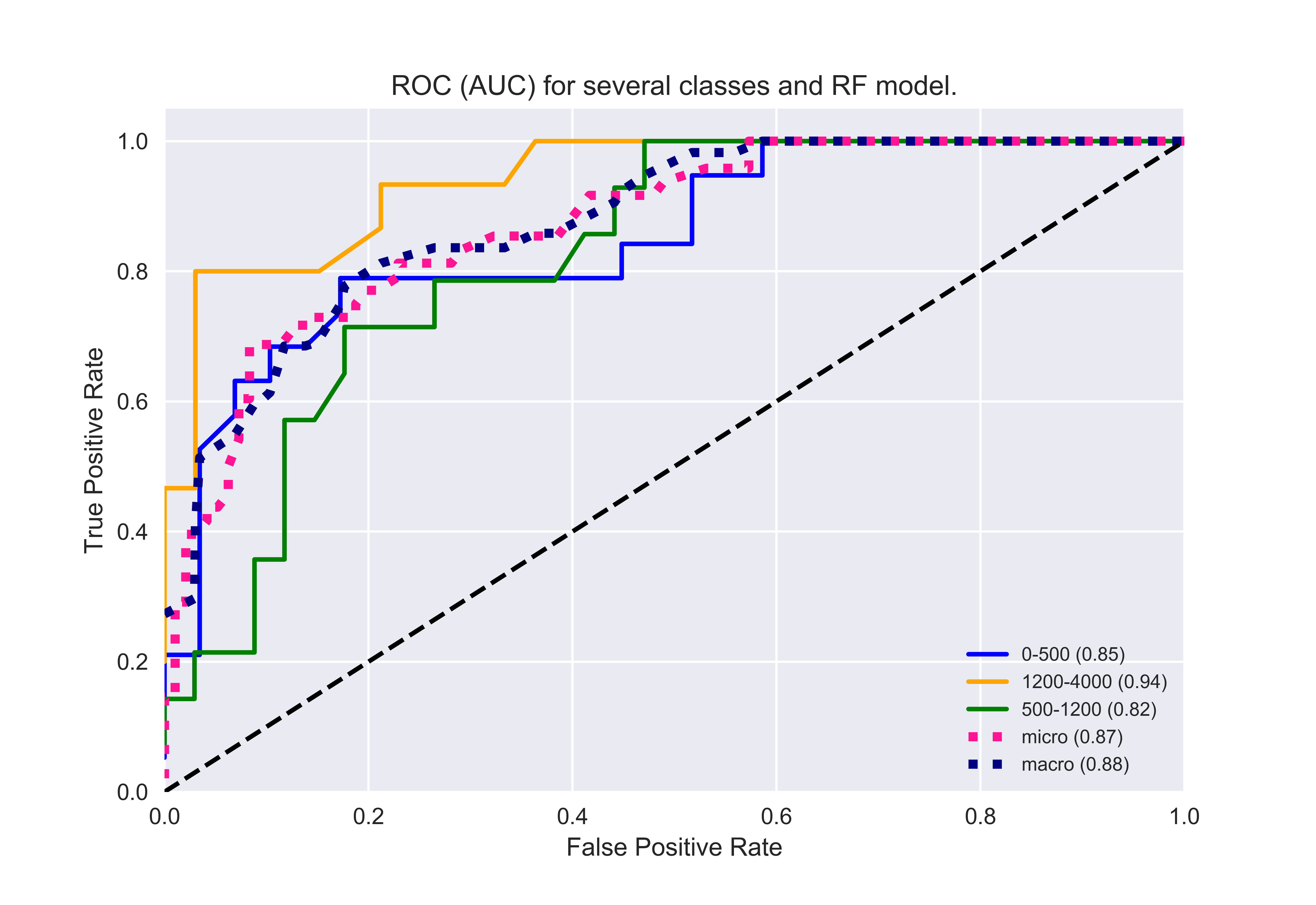}
    \caption{Receiver operating characteristic (ROC) curves and area under ROC-curves (AUC) for \emph{post}-exercise prediction for RF method (for separate classes and for the whole dataset).}
    \label{fig7_POST_05_ROC_several_classes_RF}
\end{figure}

In this case, the best performance is demonstrated by the RF classifier ($AUC_{micro}=0.87$ and $AUC_{macro}=0.88$), the DT calssifier ($AUC_{micro}=0.81$ and $AUC_{macro}=0.82$), and the KN classifier ($AUC_{micro}=0.80$ and $AUC_{macro}=0.82$).
For the RF method, the ROC-curves with AUC-values for different classes (levels of training load in calories) and for the whole dataset are shown in Fig.\ref{fig7_POST_05_ROC_several_classes_RF}. 

\begin{figure}[!hbt]
    \centering
    \includegraphics[trim={1cm 2cm 2cm 2cm},clip,width=7cm]{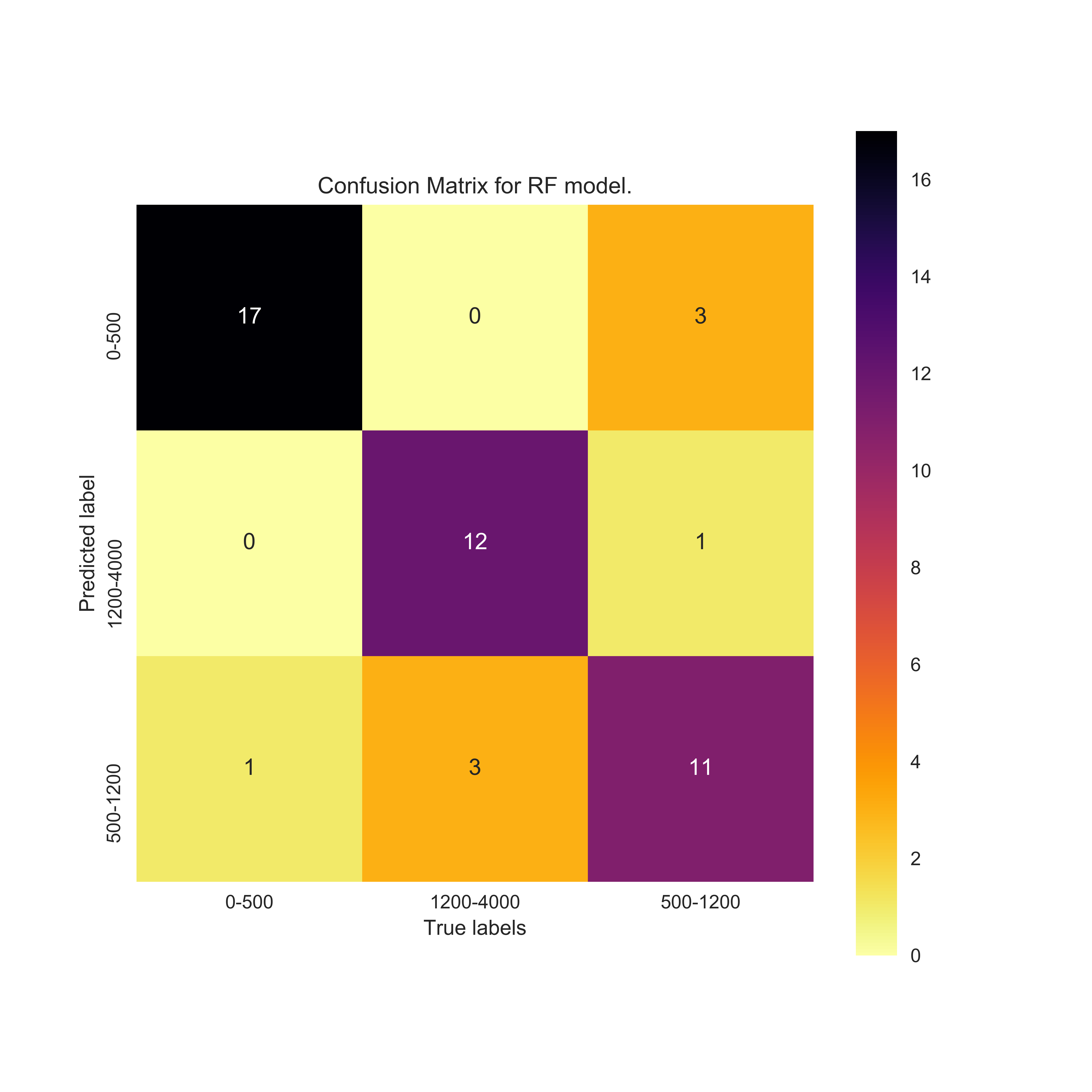}
    \caption{Confusion matrix for \emph{post}-exercise prediction by RF method.}
    \label{fig8_POST_03_confusion_matrix_RF}
\end{figure}

The confusion matrix for prediction of training load levels by RF method by the parameters recorded during the training for the test data (25\% of the whole dataset) is shown on Fig.\ref{fig8_POST_03_confusion_matrix_RF}.

\section{Discussion}\label{Discussion}
The results of in-exercise prediction of training load levels are in the good agreement with the load values in calories reported by manufacturer of the professional device used for the experiments despite their exact algorithms are proprietary and unknown. The known details are described elsewhere and reported that neural network was used for calibration and estimation of in-exercise training load values in calories \cite{Firstbeat_2015, Rennie_2001, Hiilloskorpi_2003, Pulkkinen_2005, Vesterinen_2013}. This experiment was necessary to test the proposed methodology of energy expenditure estimation by means of various ML methods and determine the most appropriate among them.

Despite the significantly lower duration of post-exercise measurements (1 minute) in comparison to in-exercise measurements (from 15 minutes to 9 hours), the prediction models demonstrate the quite high performance by k-fold cross validation tests (Fig.\ref{fig5_POST_01_model_comparison} and Table.\ref{table2_POST_01_model_comparison}), ROC-curves (Fig.\ref{fig6_POST_06_ROC_model_comparison}), and confusion matrix (Fig.\ref{fig8_POST_03_confusion_matrix_RF}). 

It is possible to improve the results by exclusion of some features from the initial model (\ref{eq_post_model_full}), where all heart activity parameters (AVNN, SDNN, RMSSD, NN50, pNN50, HRV, RAHR, RMHR) and type of activity ($A$) are used. For example, the shorter model (\ref{eq_post_model_short}) with only 4 the most important heart activity parameters ($AVNN$, $SDNN$, $RMSSD$, $NN50$, $pNN50$, $HRV$, $RAHR$, $RMHR$) and type of activity ($A$) can be used. 

\begin{equation}
    \begin{split}
	    C \sim A+AVNN+SDNN+RMSSD+HRV 
    \end{split}	
	\label{eq_post_model_short}
\end{equation}

As a result of experiments, the shorter model (\ref{eq_post_model_short}) allows us to get the better predictions with $AUC_{micro}=0.91$ and $AUC_{macro}=0.89$ for the k-nearest neighbors classifier.

It should be noted that exclusion of the type of activity ($A$) from the models (\ref{eq_in_model_full})-(\ref{eq_post_model_short}) radically decreases their performance and does not allow to make the reliable predictions on the levels of in-exercise training load. These results are compared in Table \ref{table3_POST_woType_comparison}, where column ``ML method'' contains the ML method names,  ``Model (eq.)'' --- the references on the models, columns ``Acc'' --- the mean values (column ``mean'') and the standard deviation values (column ``std'') of accuracy, column ``Prec'' --- the precision values, and column ``Rec'' --- the recall values for post-exercise prediction with the type of activity ($+A$) and without it ($-A$). This difference can be explained by the significantly different patterns of energy expenditure and reaction of cardiovascular system in very different types of activities like swimming, cycling, and running. 

\begin{table}[!hb]
    \caption{Comparison of AUC, accuracy, precision, and recall values for ML methods and models for post-exercise prediction with and without the type of activity $A$.}
    \label{table3_POST_woType_comparison}
    \centering
    \begin{tabular}{|c|c|c|c|c|c|c|c|}
        \hline
        ML & Model & AUC & AUC & Acc & Acc & Prec & Rec\\
        method & (eq.) & micro & macro & mean & std &  & \\
        \hline
        \hline
        RF & (\ref{eq_post_model_full})$+A$ & 0.87 & 0.88 & 0.65 & 0.07 & 0.84 & 0.83\\
        \hline
        RF & (\ref{eq_post_model_full})$-A$ & 0.67 & 0.65 & 0.48 & 0.04 & 0.50 & 0.48\\
        \hline
        KN & (\ref{eq_post_model_short})$+A$ & 0.91 & 0.89 & 0.74 & 0.09 & 0.85 & 0.85\\
        \hline
        KN & (\ref{eq_post_model_short})$-A$ & 0.60 & 0.60 & 0.44 & 0.09 & 0.39 & 0.37\\
        \hline
    \end{tabular}
\end{table}

Comparison of these results for in-exercise and post-exercise predictions allow to state that the proposed ML methods can bring to the light some correlation between the actual in-exercise training load and subsequent physical state of the athlete measured by the athlete's heart rate activity.

The obtained predictions cannot be generalized at the moment, because they can be very dependent on many additional aspects (for example, personal characterstics as an gender, age, weight, fitness, mental state, accumulated fatigue, etc.) of persons under tests. But from the practical point of view they can be useful for personal usage to characterize the actually obtained training loads in three categories at least after training the ML models on the personalized user data. The data and results obtained and presented here allow us to widen the range of application of these methods for the more precise prediction under condition of taking into account the time of measurements and time intervals between trainings and post-exercise measurements. In this connection, the more complicated models should be used like neural networks (NN) and deep learning (DL) neural networks, including recurrent DL NNs. For example, the models presented here can be used to estimate the current and aggregated physical training load. By estimation of the decrease of correlation between the actual in-exercise training load and post-exercise training load one can make conclusions on the recovery abilities of the person under investigation and related fatigue. It can allow us to predict the potential dangerous development with online cautions and warnings, that is very important for many health and elderly care applications \cite{Gordienko_elderly_2017,Gang_2018}. The silimalr approaches had proved to be efficient and wide-spread now in various fields like automated detection of physiological response \cite{Gang_2018, 2017_Gogate_deception, 2018_Gogate_speech} and its variations. But for this purpose additional research with usage of much larger datasets will be crucially important. In this connection, the more promicing progress can be obtained by creating and sharing the similar datasets around the world in the spirit of open science, volunteer data collection, processing and computing \cite{Gordienko_volunteer_2015, Goldberger_2000, Chen_2017}. 

\section{Conclusions}\label{Conclusions}
This work reports about application of some machine learning methods (linear regression, linear discriminant analysis, k-nearest neighbors, decision tree, random forest, Gaussian naive Bayes, support-vector machine) for monitoring energy expenditures in athletes. The  classification problem was to predict the known level of the in-exercise loads (in three categories by calories) by the heart rate activity features measured during the short period of time (1 minute only) after training, i.e by features of the post-exercise load.
The results obtained shown that the post-exercise heart activity features preserve the information of the in-exercise training loads and allow us to predict their actual in-exercise levels. The best performance can be obtained by the random forest classifier with all 8 heart rate features ($AUC_{micro}=0.87$ and $AUC_{macro}=0.88$) and the k-nearest neighbors classifier with 4 most important heart rate features ($AUC_{micro}=0.91$ and $AUC_{macro}=0.89$).
The limitations of the methods proposed consist in the limited scale of the data, absence of results as to decay of the parameters with time, and inability to predict the absolute values of energy expenditures. But these drawbacks can be excluded by increase of the dataset and improving predictions by the current DL models, including recurrent ones that are the topics of the current and future investigations.


\section*{Acknowledgment}
The work was partially supported by Huizhou Science and
Technology Bureau and Huizhou University (Huizhou,
P.R.China) in the framework of Platform Construction for
China-Ukraine Hi-Tech Park.



%\bibliographystyle{IEEEtran}

\end{document}